\documentclass{article} 
\usepackage{iclr2018_conference,times}
\usepackage{hyperref}
\usepackage{url}
\usepackage[utf8]{inputenc} 
\usepackage[T1]{fontenc}    

\usepackage{booktabs}       
\usepackage{amsfonts}       
\usepackage{nicefrac}       
\usepackage{microtype}      

\usepackage{natbib}
\usepackage[]{algorithm2e}
\usepackage{graphicx}
\usepackage{amsmath}
\usepackage{spverbatim}

\usepackage{xcolor}

\newcommand{\rmsnorm}{\mathit{RMS}_\mathrm{\kern-0.5pt norm}}
\newcommand{\MSg}{\mathit{MS}_{\kern-1.2pt g}}

\usepackage[colorinlistoftodos]{todonotes}
\usepackage{hyperref}
\usepackage{subcaption}
\usepackage{multirow}
\usepackage{adjustbox}

\title{\scalebox{0.85}{Dynamic Evaluation of Transformer Language Models}}

\author{Ben Krause, Emmanuel Kahembwe, Iain Murray, \& Steve Renals \\
School of Informatics, University of Edinburgh \\
Edinburgh, Scotland, UK \\
\texttt{ben.krause,e.kahembwe,i.murray,s.renals@ed.ac.uk} \\
}

\iclrfinalcopy 

\begin{document}

\maketitle

\begin{abstract}
This research note combines two methods that have recently improved the state of the art in language modeling: Transformers and dynamic evaluation. Transformers use stacked layers of self-attention that allow them to capture long range dependencies in sequential data. Dynamic evaluation fits models to the recent sequence history, allowing them to assign higher probabilities to re-occurring sequential patterns. By applying dynamic evaluation to Transformer-XL models, we improve the state of the art on enwik8 from 0.99 to 0.94 bits/char, text8 from 1.08 to 1.04 bits/char, and WikiText-103 from 18.3 to 16.4 perplexity points. Code to replicate our results is available\footnote{\url{www.github.com/benkrause/dynamiceval-transformer}}.
\end{abstract}

\section{Introduction}

Language modeling is a commonly used machine learning benchmark with applications to speech recognition, machine translation, text generation, and unsupervised learning in natural language processing tasks. LSTMs \citep{Hochreiter-1997} conventionally used for language modeling have been shown to use relatively shorter contexts to make predictions \citep{khandelwal2018}. Several recent improvements to language modeling have resulted from models with increased ability to use long-range dependencies. This work combines two specific advances: Transformers \citep{vaswani2017} and dynamic evaluation \citep{mikolov2010,krause2018}. Transformers can model long range dependencies through stacked layers of self-attention, and dynamic evaluation exploits certain types of long range dependencies by adapting parameters based on the observed sequence history. Dynamic evaluation can be applied to any language model at test time, but to our knowledge, no previous work has applied dynamic evaluation to Transformers. 

 Transformers use a combination of a self-attention mechanism and positional embeddings to encode information about the sequence history \citep{vaswani2017}. The use of self-attention provides shorter paths for information to travel, which is conjectured to be one of the main reasons that transformers achieve better results on common language modeling benchmarks compared to other models \citep{dai2019}. Moreover, transformers trained on very large datasets can generalize to other NLP tasks, and generate realistic samples that are coherent over long time frames \citep{radford2019}.

Dynamic evaluation adapts models to the recent sequence history via gradient descent in order to exploit re-occurring sequential patterns. Natural language tends to have long range dependencies associated with the style and word usage of particular passages of texts; and dynamic evaluation can exploit these dependencies via online model adaptation. Transformers with a large memory cache also potentially have the capability of adapting to the style of the recent sequence history, although it is unclear to what extent they learn to do this in practice. Dynamic evaluation and Transformers have each shown their respective capabilities to use thousands of timesteps of context to improve predictions \citep{krause2018,dai2019}, but it is unclear how much overlap there is between the type of long-range dependencies exploited by Transformers and dynamic evaluation. If Transformers are able to fully adapt to the style of the recent sequence history, there should be little to no advantage of using dynamic evaluation. Therefore, in this work, we explore the utility of applying dynamic evaluation to Transformers.

\section{Transformers}
\label{sec:transformer}

A number of variants of Transformers have been suggested for language modeling \citep{al2018,liu2018,baevski2018,radford2018}, but in this work, we focus on the Transformer-XL architecture of \citet{dai2019}, which uses segment-level attention recurrence and a relative positional encoding mechanism to generalize to longer attention lengths than seen during training. Transformer-XL has recently improved  state-of-the-art results on a number of common language modeling benchmarks.

The Transformer-XL, like the regular Transformer, contains stacked 
self-attention layers and position-wise feedforward operations. The Transformer-XL processes sequence segments in parallel across time in each forward pass. The hidden states from these sequence segments are cached in a memory so that future sequence segments can apply attention over them. We refer to  \citet{dai2019} for the full details of the model.

\section{Dynamic evaluation}
\label{sec:dyneval}

Dynamic evaluation is a gradient descent based adaptation method that can be applied to auto-regressive sequence modeling problems. Auto-regressive sequence models use the following factorization to assign a probability to a sequence $x_{1:T} = \{x_1,\dots,x_T\}$:

\begin{equation}
P(x_{1:T}) = P(x_1) \prod_{t=2}^{T} P(x_t|x_{1:t-1}).
\end{equation}

\begin{figure}[tb]
  \centering
  \includegraphics[width=\columnwidth]{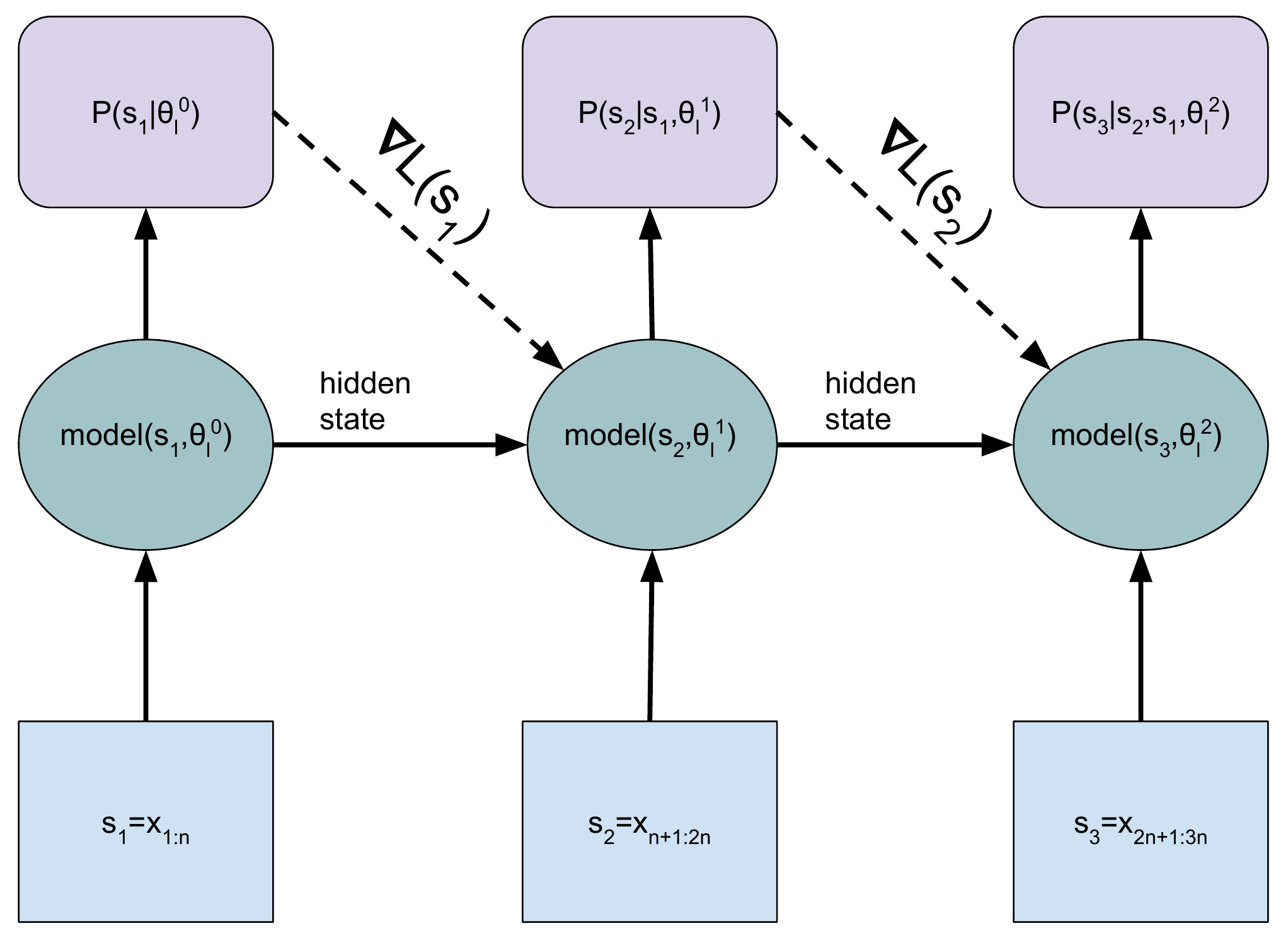}
  \caption{Illustration of dynamic evaluation (figure from \citet{krause2018}). The model evaluates the probability of sequence segments $s_i$ of length $n$. The gradient $\nabla\mathcal{L}(s_i)$ with respect to the log probability of $s_i$ is used to update the model parameters $\theta^{i-1}_l$ to $\theta^{i}_l$ before the model progresses to the next sequence segment. Dashed edges are what distinguish dynamic evaluation from static (normal) evaluation. }
  \label{fig:dynamiceval}
\end{figure}

The model predicts a distribution over the next sequence element $P(x_t|x_{1:t-1})$ (or a sequence segment $x_{t_1:t_2}|x_{1:t_1-1}$).  The model observes the true $x_t$ and takes a loss based on the cross entropy prediction error, $\mathcal{L}_t$. The gradient $\nabla\mathcal{L}_t$ is then used to update the network before proceeding to the next sequence element. As in all autoregressive models, dynamic evaluation only conditions on sequence elements that it has already predicted, and so evaluates a valid log-probability for each sequence. Dynamic evaluation is illustrated graphically in figure \ref{fig:dynamiceval}.

The gradient descent adjusted weights can be interpreted as a memory that can better capture re-occurring patterns that occur in linguistic sequences. Dynamic evaluation updates were shown to have the ability to increase probabilities of words that occur in a sequence, as well as words with similar embeddings to words that occur in the sequence \citep{krause2018}. This capability gives dynamic evaluation the potential to better model recently seen words, as well as to adapt more broadly to the style and topic of a sequence.

Following \citet{krause2018}, which applies dynamic evaluation to RNNs at the sequence segment level, we apply dynamic evaluation to Transformer-XL models at the sequence segment level. 
Since Transformer-XL models are designed to process sequences in segments, we align the sequence segments used for Transformer-XL \citep{dai2019} with the sequence segments used to compute the gradient for dynamic evaluation. The gradient is computed once for each sequence segment (after taking a loss on the segment), and backpropagation is truncated to be contained within a single sequence segment.

There is a large space of potential optimizers that can be used for dynamic evaluation, and we evaluate two in this work. We consider a simple baseline that uses stochastic gradient descent with a fixed learning rate to update the weights of the network on each segment. We also consider the more complex dynamic evaluation optimizer \citep{krause2018} which uses an update rule related to RMSprop \citep{tieleman2012}, except that gradient statistics are computed from the training data, and weights are decayed back to the original parameters learned during training.

\section{Experiments}

We applied dynamic evaluation to pretrained Transformer-XL models from \citet{dai2019} on two character-level datasets and one word-level dataset. We chose these 3 datasets because they all contain long-range dependencies that span across sentences and paragraphs. Details of the model training can be found in \citet{dai2019}, and we downloaded models using their code\footnote{\url{https://github.com/kimiyoung/transformer-xl}}.

We measured the performance of two types of dynamic evaluation; one which used the optimizer from \citet{krause2018}, which we refer to as ``RMS dynamic eval + decay'', and one that used stochastic gradient descent, which we refer to as ``SGD dynamic eval''. Following \citet{krause2018}, we tuned hyperparameters for dynamic evaluation on the validation sets before evaluating on the test sets.

\subsection{Character-level experiments}
\label{sec:clexpt}

We use two datasets to evaluate dynamic evaluation on character-level Transformer-XL models; enwik8 \citep{Hutter-2006} and text8\footnote{\url{http://mattmahoney.net/dc/textdata}}. enwik8 is a byte-level data set derived from Wikipedia that in addition to English text, also includes markup, special characters, and text in other languages. enwik8 contains 90M characters for training, 5M for validation, and 5M for testing. We noticed a slight anomaly in the preprocessing of enwik8 in the code released by \citet{dai2019} that caused it to have 204 unique tokens (rather than the standard 205 tokens used in most results, for instance in \citet{Graves-2013}), and our results also contain this anomaly since we use pretrained models from their work. text8 is derived from the same data as enwik8, but is preprocessed to only contain an alphabet of 27 characters (lowercase a--z plus spaces). text8 also uses a 90M--5M--5M split for training, validation, and testing. Following \citet{dai2019}, we used sequence segments of 128 and a memory cache of length 3800 for both datasets. Results for enwik8 and text8 are reported in Table~\ref{tab:dynamic-transformer-char} and Table~\ref{tab:dynamic-transformer-text8} respectively. Applying Dynamic evaluation improves the Transformer-XL by a noticeable margin, achieving state of the art on both of these character-level datasets.

\begin{table*}[tb]
\begin{center}
\begin{adjustbox}{width=1\textwidth}
\begin{tabular}{l  l  l  } \toprule
Model & \# of params  & test   \\
\midrule
Hyper LSTM \citep{Ha2017}  & 25M & 1.34 \\
HM-LSTM \citep{chung2017}  & 35M & 1.32 \\
Recurrent highway networks \citep{zilly2017} & 46M & 1.27   \\
FS-LSTM \citep{mujika2017} & 47M & 1.25 \\
awd-LSTM \citep{merity2018} & 47M & 1.23 \\
Transformer + aux losses \citep{al2018}  & 235M & 1.06 \\
Multiplicative LSTM \citep{krause2016} & 46M & 1.24  \\
Multiplicative LSTM + dynamic eval \citep{krause2018} & 46M & 1.08 \\
\midrule
Transformer-XL \citep{dai2019} & 277M  & 0.993 \\
\textbf{Transformer-XL + SGD dynamic eval} & 277M  & 0.946   \\
\textbf{Transformer-XL + RMS dynamic eval + decay} & 277M  & \textbf{0.940}     \\
\bottomrule
\end{tabular}
\end{adjustbox}
\end{center}
\caption{Character-level cross-entropy (bits/char) on enwik8. As noted in
Section~\ref{sec:clexpt}, there is a slight difference in the data used in the
final three results and previous work.}
\label{tab:dynamic-transformer-char}
\end{table*}
\begin{table*}[tb]
\begin{center}
\begin{adjustbox}{width=1\textwidth}
\begin{tabular}{l  l  l  } \toprule
Model & \# of params & test \\
\midrule
HM-LSTM \citep{chung2017} & 35M & 1.29 \\
Recurrent highway networks \citep{zilly2017} & 45M & 1.27   \\
Transformer + aux losses \citep{al2018} & 235M & 1.13 \\
Multiplicative LSTM \citep{krause2016} & 45M & 1.27  \\
Multiplicative LSTM + dynamic eval \citep{krause2018} & 45M & 1.19 \\
\midrule
Transformer-XL \citep{dai2019} & 277M  & 1.085 \\
\textbf{Transformer-XL + SGD dynamic eval} &  277M  & 1.042  \\
\textbf{Transformer-XL + RMS dynamic eval + decay} & 277M  & \textbf{1.038}   \\
\bottomrule
\end{tabular}
\end{adjustbox}
\end{center}
\caption{Character-level cross-entropy (bits/char) on text8.}
\label{tab:dynamic-transformer-text8}
\end{table*}

\subsection{Word-level experiments}

We evaluate dynamic evaluation on word-level Transformer-XL using the WikiText-103 dataset \citep{Merity2016}, which is also comprised of Wikipedia text. WikiText-103 contains 103 million training tokens, and a vocabulary size of 268k. Given the large vocabulary size, the pretrained model we re-evaluate from \citet{dai2019} used an adaptive softmax output layer \citep{grave2017a} to make training faster. Results for WikiText-103 are reported in Table~\ref{tab:dynamic-transformer-word}. There was no noticeable validation advantage to using a decay rate, so we refer to the dynamic evaluation optimizer for this experiment simply as ``RMS dynamic eval'', since the decay rate was tuned to be set to zero. Dynamic evaluation gave a 9\% perplexity improvement to Transformer-XL on WikiText-103.

\begin{table*}[tb]
\begin{center}
\begin{adjustbox}{width=1\textwidth}
\begin{tabular}{l  l  l  l} \toprule
Model & \# of params & valid & test  \\
\midrule
LSTM+ neural cache \citep{grave2017b} & - & - & 40.8 \\
GCNN-14 \citep{dauphin2017} & - & - & 37.2 \\
QRNN \citep{merity2018} & 151M  & 32.0 & 33.0  \\
LSTM + hebbian + cache \citep{rae2018} & - & 29.7 & 29.9  \\
Transformer + adaptive input \citep{baevski2018}  & 247M & 19.8 & 20.5 \\
\midrule
Transformer-XL${}^*$ \citep{dai2019}  & 257M & 17.3 & 18.1 \\
\textbf{Transformer-XL + SGD dynamic eval}  & 257M & 16.3 & 17.0 \\
\textbf{Transformer-XL + RMS dynamic eval }  & 257M & 15.8 & \textbf{16.4}  \\
\bottomrule
\end{tabular}
\end{adjustbox}
\end{center}
\caption{Word-level perplexity on WikiText-103. ${}^*$We report our results using the pretrained model from \citet{dai2019} using a batch size of 1, and achieved a slightly lower perplexity than in the original paper (18.1 vs 18.3).}
\label{tab:dynamic-transformer-word}
\end{table*}

The results on WikiText-103 are the first that we know of that apply dynamic evaluation with an adaptive softmax output layer. Adaptive softmax reduces the computational expense of the output layer at the cost of giving the model less expressiveness at modeling rare words. When training a network from scratch, such a trade-off is sensible, since it is difficult to learn a good representation of rare words. However, when dynamically adapting to the recent sequence history, the adaptive softmax layer may make adapting to recent rare words more challenging. There is potential for future work improving the combination of dynamic evaluation and adaptive softmax, for instance by hybridizing it with the neural cache method \citep{grave2017b}. The neural cache learns a non-parametric output layer that is independent of the network's output layer, which may potentially allow for more expressive adaptation to rare words in models with an adaptive softmax.

\section{Conclusion}

Dynamic evaluation was able to give moderate improvements to strong Transformer network baselines, and improves the state of the art on all three datasets evaluated. These results demonstrate that the types of long range dependencies used by dynamic evaluation and Transformers are somewhat different, as applying dynamic evaluation to Transformers leads to further improvements. These improvements are not nearly as large as when dynamic evaluation has been applied to weaker models, suggesting that Transformers are by themselves somewhat more capable of modeling re-occurring patterns in sequences than past architectures. However, Transformers still struggle to fully exploit these repetitions, even in these experiments where training and testing data came from the same domain. Transformers may struggle to adapt even more when there is a shift between training and testing data. Our results therefore motivate future work on enhancements and architectures for adaptive sequence modeling, as current Transformer models cannot fully deal with adaptation on their own.

\bibliographystyle{apalike}
\bibliography{iclr2018_conference}
\end{document}